%% file: main.tex
\documentclass[a4paper,conference]{IEEEtran}
\usepackage{bigdelim}
\usepackage{rotating}
\usepackage[utf8]{inputenc} % allow utf-8 input
\usepackage[T1]{fontenc}    % use 8-bit T1 fonts
\usepackage{hyperref}       % hyperlinks
\usepackage{url}            % simple URL typesetting
\usepackage{booktabs}       % professional-quality tables
\usepackage{amsfonts}       % blackboard math symbols
\usepackage{nicefrac}       % compact symbols for 1/2, etc.
\usepackage{microtype}      % microtypography
\usepackage{xcolor}         % colors
\usepackage{tikzit}
\usepackage{tikz}
\usepackage{array,multirow}
\def\checkmark{\tikz\fill[scale=0.4](0,.35) -- (.25,0) -- (1,.7) -- (.25,.15) -- cycle;} 
\input{tikz/sample.tikzstyles}
\usepackage{nidanfloat} % for table placement in two column mode
\usepackage{lipsum}  
\usepackage[pangram]{blindtext}
\usepackage{placeins}
\usepackage{comment}

\usepackage{pgfplots}

\makeatletter
\def\ps@myheadings{%
    \let\@oddfoot\@empty\let\@evenfoot\@empty
    \def\@evenhead{\thepage\hfil\slshape\leftmark}%
    \def\@oddhead{{\slshape\rightmark}\hfil\thepage}%
    \let\@mkboth\@gobbletwo
    \let\sectionmark\@gobble
    \let\subsectionmark\@gobble
    }
  \if@titlepage
  \renewcommand\maketitle{\begin{titlepage}%
  \let\footnotesize\small
  \let\footnoterule\relax
  \let \footnote \thanks
  \null\vfil
  \vskip 60\p@
  \begin{center}%
    {\LARGE \@title \par}%
    \vskip 3em%
    {\large
     \lineskip .75em%
      \begin{tabular}[t]{c}%
        \@author
      \end{tabular}\par}%
      \vskip 1.5em%
    {\large \@date \par}%       % Set date in \large size.
  \end{center}\par
  \@thanks
  \vfil\null
  \end{titlepage}%
  \setcounter{footnote}{0}%
}
\else
\renewcommand\maketitle{\par
  \begingroup
    \renewcommand\thefootnote{\@fnsymbol\c@footnote}%
    \def\@makefnmark{\rlap{\@textsuperscript{\normalfont\@thefnmark}}}%
    \long\def\@makefntext##1{\parindent 1em\noindent
            \hb@xt@1.8em{%
                \hss\@textsuperscript{\normalfont\@thefnmark}}##1}%
    \if@twocolumn
      \ifnum \col@number=\@ne
        \@maketitle
      \else
        \twocolumn[\@maketitle]%
      \fi
    \else
      \newpage
      \global\@topnum\z@   % Prevents figures from going at top of page.
      \@maketitle
    \fi
    \thispagestyle{plain}\@thanks
  \endgroup
  \setcounter{footnote}{0}%
}
\makeatother

\title{\scalebox{0.93}{EASY -- Ensemble Augmented-Shot Y-shaped Learning:}\\\scalebox{0.78}{State-Of-The-Art Few-Shot Classification with Simple Ingredients}}

\author{\IEEEauthorblockN{Yassir Bendou\IEEEauthorrefmark{1},
Yuqing Hu\IEEEauthorrefmark{1}\IEEEauthorrefmark{2},
Raphael Lafargue\IEEEauthorrefmark{1},
Giulia Lioi\IEEEauthorrefmark{1},
Bastien Pasdeloup\IEEEauthorrefmark{1}\\ 
Stéphane Pateux\IEEEauthorrefmark{2} and 
Vincent Gripon\IEEEauthorrefmark{1}}
\IEEEauthorblockA{\IEEEauthorrefmark{1}IMT Atlantique, Technopole Brest Iroise, France}
\IEEEauthorblockA{\IEEEauthorrefmark{2}Orange Labs, Rennes, France}}

\usepackage{amsmath}

\begin{document}

\maketitle

\begin{abstract}
Few-shot learning aims at leveraging knowledge learned by one or more deep learning models, in order to obtain good classification performance on new problems, where only a few labeled samples per class are available. Recent years have seen a fair number of works in the field, introducing methods with numerous ingredients. A frequent problem, though, is the use of suboptimally trained models to extract knowledge, leading to interrogations on whether proposed approaches bring gains compared to using better initial models without the introduced ingredients. In this work, we propose a simple methodology, that reaches or even beats state of the art performance on multiple standardized benchmarks of the field, while adding almost no hyperparameters or parameters to those used for training the initial deep learning models on the generic dataset. This methodology offers a new baseline on which to propose (and fairly compare) new techniques or adapt existing ones.
\end{abstract}

%\begin{itemize}
%    \item finaliser related work, rajouter plusieurs refs
%    \item finaliser les tables de comparaison avec état de l'art, remplir les tables en transductif
%   \item table 11, rajouter discussion sur l'ensembliste (quand 3-miniResnet marche moins qu'un grand Resnet car le dataset est grand)
% \item transductif : tester avec soft k-means cosine similarity des centroides 
%\end{itemize}

\IEEEpeerreviewmaketitle

\section{Introduction}

Learning with few examples, or \emph{few-shot learning}, is a domain of research that has become increasingly popular in the past few years. Reconciling the remarkable performances of deep learning (DL), which are generally obtained thanks to access to huge databases, with the constraint of having a very small number of examples may seem paradoxical. Yet the answer lies in the ability of DL to transfer knowledge acquired when solving a previous task toward a different new one.

The classical few-shot setting consists of two parts:
\begin{itemize}
    \item A \emph{generic dataset}, which contains many examples of many classes. Since this dataset does not suffer from data thriftiness, it can be used to efficiently train DL architectures. Authors often split the generic dataset into two disjoint subsets, called \emph{base} and \emph{validation}. As usual in classification, the base dataset is used during training and the validation dataset is then used as a proxy to measure generalization performance on unseen data and therefore can be leveraged to fix hyperparameters. However, contrary to common classification settings, in few-shot the validation and base datasets usually contain distinct classes, so that the generalization performance is assessed on new classes. Drawing knowledge from the generic dataset can be performed with multiple strategies, as will be further discussed in Section~\ref{sec:rw};
    
    \item A \emph{novel dataset}, which consists of classes that are distinct from those of the generic dataset. We are only given a few labeled examples for each class, resulting in a few-shot problem. The labeled samples are often called the \emph{support} set, and the remaining ones the \emph{query} set. When benchmarking, it is common to use a large novel dataset from which artificial few-shot tasks are sampled uniformly at random, what we call a \emph{run}. In that case, the number of classes $n$ (named \emph{ways}), the number of shots per class $k$ and the number of query samples per class $q$ are given by the benchmark. Reported performance are often averaged over a large number of runs.
    
\end{itemize}

In order to exploit knowledge previously learned by models on the generic dataset, a common approach is to remove their final classification layer. The resulting models, now seen as feature extractors, are generally termed \emph{backbones}, and can be used to transform the support and query datasets into \emph{feature vectors}. This is a form of transfer learning. In this work, we do not consider the use of additional data such as other datasets, neither semantic nor segmentation information. Additional preprocessing steps may also be used on the samples and/or on the associated feature vectors, before the classification task. Another major approach uses meta-learning~\cite{Finn2017, Munkhdalai2018, Lee, Scott, Munkhdalai, Zhang}, as mentioned in Section \ref{sec:rw}.

It is important to distinguish two types of problems:
\begin{itemize}
    \item In \emph{inductive} few-shot learning, only the support dataset is available to the few-shot classifier, and prediction is performed on each sample of the \emph{query} dataset independently from each other;
    \item In \emph{transductive} few-shot learning, the few-shot classifier has access to both the support and the full query datasets when performing predictions.
\end{itemize}
Both problems have connections with real-world situations. In general, inductive few-shot corresponds to cases where data acquisition is expensive, whereas transductive few-shot corresponds to cases where data labeling is expensive.

\begin{figure*}
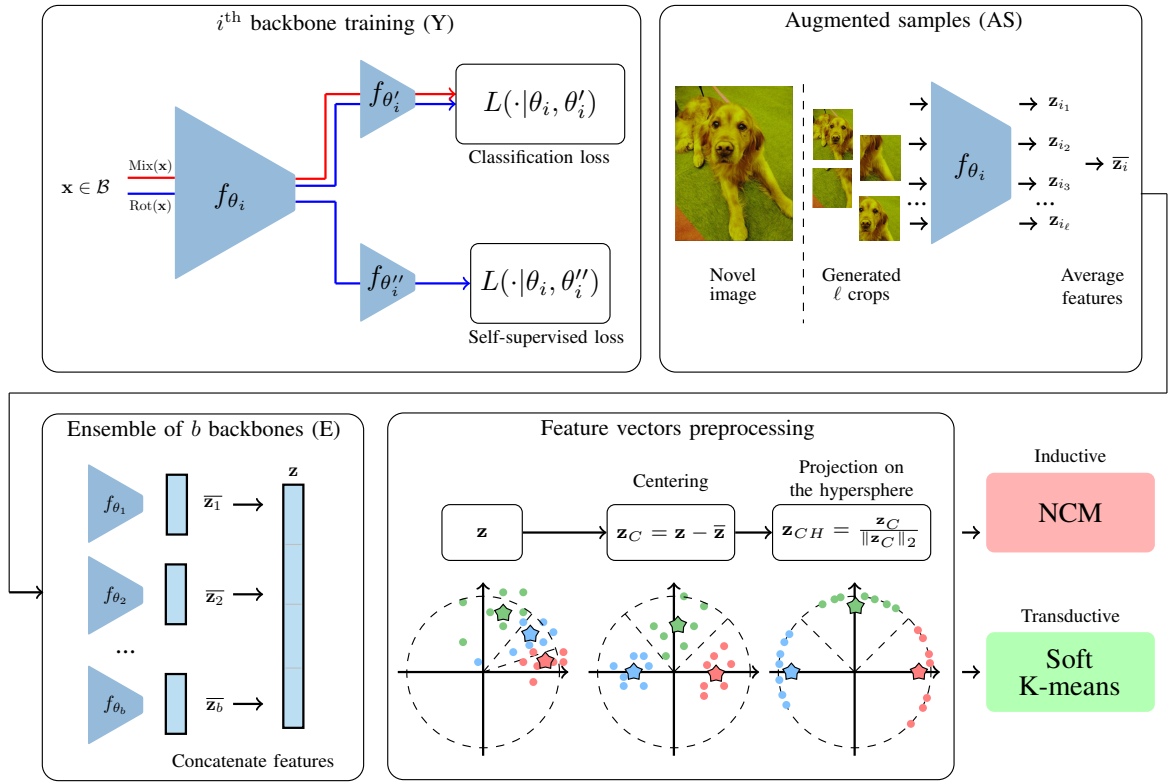


\ctikzfig{tikz/fig1}
\caption{Illustration of our proposed method. \textbf{Y}: We first train an ensemble of backbones using the generic dataset. We use two cross-entropy losses in parallel: one for the classification of base classes, and the other for the self-supervised targets (rotations). We also use manifold mixup~\cite{Verma}. All the backbones are trained using the exact same routine, except that their initialization is different (random) and the order in which data batches are presented is also potentially different; \textbf{AS}: Then, for each image in the novel dataset, we generate multiple crops, then compute their feature vectors, that we average; \textbf{E}: Each image becomes represented as the concatenation of the outputs of AS for each backbone; \textbf{Preprocessing}: We add a few classical preprocessing steps, including centering by removing the mean of the feature vectors of the base dataset in the inductive case or the novel feature vectors for the transductive case, and projecting on the hypersphere. Finally, we use a simple nearest class mean classifier (NCM) if in inductive setting or a soft K-means algorithm in transductive setting.} \label{fig:method}
\end{figure*}

In recent years, a large number of contributions have introduced methodologies to cope with few-shot problems. There are a lot of ingredients involved, including distillation~\cite{ma2021partner}, contrastive learning~\cite{luo2021rectifying}, episodic training~\cite{Snell2017}, mixup ~\cite{Zhang2018}, manifold mixup~\cite{mangla2020charting, Verma} and self-supervision~\cite{mangla2020charting}. As a consequence, it can appear quite opaque what are the effective ingredients, and whether their performance can be reproduced across different datasets or settings. More problematically, we noticed that many of these contributions start with suboptimal training procedures or architectures. Admittedly, they show significant performance boost using their proposed method, but reach at the end only a fair performance compared with better initial models without the proposed ingredient.

In this paper, we are interested in proposing a very simple method combining ingredients commonly found in the literature and yet achieving competitive performance. As such this contribution does not propose anything completely new, but we believe it will help having a clearer view on how to efficiently implement few-shot learning for real-world applications. Our main motivation is to define a proper baseline to compare to and to start with, on which obtaining boost of performance is going to be much more challenging than starting from a poorly trained backbone. We also aim at showing that a simple approach reaches higher performance than increasingly complex methods proposed in the recent few-shot literature.

More precisely, in this paper, we:
\begin{itemize}
    \item Introduce a very simple methodology, illustrated in Figure \ref{fig:method}, for inductive or transductive few-shot learning, that comes with almost no hyperparameters but those used for training the backbones;
    \item Show the ability of the proposed methodology to reach or even beat state-of-the-art performance on multiple standardized benchmarks of the field.
\end{itemize}

\section{Related Work}
\label{sec:rw}

There have been many approaches proposed recently in the field of few-shot learning. We introduce some of them following the classical pipeline. Note that our proposed methodology uses multiple ingredients from those presented thereafter.

\subsection{Data augmentation}

First, \emph{data augmentation} or \emph{augmented sampling} are generally used on the generic dataset to artificially produce additional samples, for example using rotations~\cite{mangla2020charting}, crops~\cite{zhang2020deepemd}, jitter, GANs~\cite{choe2017face,li2020adversarial}, or other techniques~\cite{hariharan2017low}. Data augmentation on support and query sets however is less frequent. Approaches exploring this direction include \cite{luo2021rectifying}, where authors propose to select the foreground objects of images by identifying the right crops using a relatively complex mechanism; and \cite{yang2021free}, where the authors propose to mimic the neighboring base classes distribution to create augmented latent space vectors.

In addition, \emph{mixup}~\cite{Zhang2018} and \emph{manifold-mixup}~\cite{Verma} are also used to address the challenging lack of data. Both can be seen as regularization methods through linear interpolations of samples and labels. Mixup creates linear interpolations at the sample level while manifold mixup focuses on feature vectors. 

\subsection{Backbone training}

Mixup is often used in conjunction with \emph{Self-supervision} (S2)~\cite{mangla2020charting} to make backbones more robust. Most of the time, S2 is implemented as an auxiliary loss meant to train the backbone to recognize which transformation was applied to an image.

A well known training strategy is \emph{episodic training}. The idea behind it boils down to having the same train and test conditions. Thus, the backbone training strategy, often based on gradient descent, does not select random batches, but uses batches designed as few-shot problems~\cite{Finn2017, Ravi, Snell2017, NIPS2016_90e13578}.

\emph{Meta-Learning}, or \emph{learning to learn}, is a major line of research in the field. This method typically learns a good initialization or a good optimizer such that new classes can be learned in a few gradient steps~\cite{Finn2017, Munkhdalai2018, Lee, Scott, Munkhdalai, Zhang}. In this regard, episodic training is often used, and recent work leveraged this concept to generate augmented tasks in the training of the backbone~\cite{liu2020task}.

\emph{Contrastive learning} aims to train a model to learn to maximize similarities between transformed instances of the same image and minimize agreement between transformed instances of different images~\cite{Luo, Liu_Fu_Xu_Yang_Li_Wang_Zhang_2021, 9428444, Majumder, luo2021rectifying}. \emph{Supervised contrastive learning} is a variant which has been recently used in few-shot learning, where similarity is maximized between instances of a class instead of the same image~\cite{NEURIPS2020_d89a66c7, ma2021partner}.

\subsection{Exploiting multiple backbones}

\emph{Distillation} has been recently used in the few-shot literature. The idea is to transfer knowledge from a teacher model to a student model by forcing the latter to match the class probabilities distribution of the teacher \cite{inproceedings, Tian, ma2021partner}.

\emph{Ensembling} consists in the concatenation of features extracted by different backbones. It was used to improve performances in few-shot learning~\cite{liu2020task}. It can be seen as a more straightforward alternative to distillation. To limit the computationally expensive training of multiple backbones, some authors propose the use of snapshots~\cite{huang2017snapshot}.
%ma2021partner

\subsection{Few-shot classification}

Over the past years, classification methods in the inductive setting have mostly relied on simple methods such as nearest class mean~\cite{wang2019simpleshot}, cosine classifiers~\cite{Chen2019} and logistic regression~\cite{yang2021free}.

More diverse methods can be implemented in the transductive setting. Clustering algorithms~\cite{luo2021rectifying}, embedding propagation~\cite{rodriguez2020embedding} and optimal transport~\cite{hu2021squeezing} were leveraged successfully to outrun performances in the inductive setting by a large margin.

\section{Methodology}

The proposed methodology consists of 5 steps, described thereafter and illustrated in Figure \ref{fig:method}. In the experiments we also report ablation results when omitting the optional steps.

\subsection{Backbone training (Y)}

We use data augmentation with random resized crops, random color jitters and random horizontal flips, which is standard in the field.

We use a cosine-annealing scheduler~\cite{loshchilov2016sgdr}, where at each step the learning rate is updated. During a cosine cycle, the learning rate evolves between $\eta_0$ and $0$. At the end of the cycle, we warm-restart the learning procedure and start over with a diminished $\eta_0$. We start with $\eta_0 = 0.1$ and reduce $\eta_0$ by 10\% at each cycle. We use 5 cycles with 100 epochs each.

We train our backbones using the methodology called S2M2R described in~\cite{mangla2020charting}. Basically, the principle is to take a standard classification architecture (\emph{e.g.}, ResNet12 ~\cite{He2016}), and to branch a new logistic regression classifier after the penultimate layer, in addition to the one used to identify the classes of input samples, thus forming a Y-shaped model~(c.f. Figure~\ref{fig:method}). This new classifier is meant to retrieve which one of four possible rotations (quarters of 360° turns) has been applied to the input samples. We use a two-step forward-backward pass at each step, where a first batch of inputs is only fed to the first classifier, combined with manifold-mixup~\cite{mangla2020charting,Verma}. A second batch of inputs is then applied arbitrary rotations and fed to both classifiers. After this training, backbones are frozen.

We experiment using a standard ResNet12 as described in~\cite{He2016}, where the feature vectors are of dimension 640. These feature vectors are obtained by computing a global average pooling over the output of the last convolution layer. Such a backbone contains $\sim 12$ million trainable parameters. We also experiment with reduced-size ResNet12, denoted ResNet12$\left(\frac{1}{2}\right)$ where we divide each number of feature maps by 2, resulting in feature vectors of dimension 320, and ResNet12$\left(\frac{1}{\sqrt{2}}\right)$, where the number of feature maps is divided roughly by $\sqrt{2}$, resulting in feature vectors of dimension 450. The numbers of parameters are respectively $\sim3$ million and $\sim6$ million.

Using common notations of the field, if we denote $\mathbf{x}$ an input sample, and $f$ the mathematical function of the backbone, then $z = f(\mathbf{x})$ denotes the feature vector associated with $\mathbf{x}$.

From this point on, we use the frozen backbones to extract feature vectors from the base, validation and novel datasets. 

\subsection{Augmented samples (AS)}

We propose to generate augmented feature vectors for each sample from the validation and novel datasets. To this end, we use random resized crops from the corresponding images. We obtain multiple versions of each feature vector and average them. In practice, we use $\ell=30$ crops per image, as larger values do not benefit accuracy much. This step is optional.

\subsection{Ensemble of backbones (E)}

To boost performance even further, we propose to concatenate the feature vectors obtained from multiple backbones trained using the same previously described routine, but with different random seeds. To perform fair comparisons, when comparing a backbone with an ensemble of $b$ backbones, we reduce the number of parameters in the ensemble backbones such that the total number of parameters remains identical. We believe that this strategy is an alternative to performing distillation, with the interest of not requiring extra-parameters and considerably reducing training time. Again, this step is optional and we perform ablation tests in the next section.

\subsection{Feature vectors preprocessing}

Finally, we apply two transforms as in~\cite{wang2019simpleshot} on feature vectors $\mathbf{z}$. Denote $\overline{\mathbf{z}}$ the average feature vector of the base dataset if in inductive setting or of the few-shot considered problem if in transductive setting.
The first operation ($C$ -- centering of $\mathbf{z}$) consists in computing:
\begin{equation}
    \mathbf{z}_C = \mathbf{z} - \overline{\mathbf{z}}\;.
\end{equation}

The second operation ($H$ -- projection of $\mathbf{z}_C$ on the hypersphere) is then:
\begin{equation}
    \mathbf{z}_{CH} = \frac{\mathbf{z}_C}{\|\mathbf{z}_C\|_2} \;.
\end{equation}

%Frow now on, we work with the feature vectors $\mathbf{z}_{ME}$ instead of $\mathbf{z}$. Note that this processing has been introduced in~\cite{}. We use the notations of~\cite{}.

\subsection{Classification}

Let us denote $\mathcal{S}_{i}$ $\left(i \in \{1, \dots, n\}\right)$ the set of feature vectors (preprocessed as $\mathbf{z}_{CH}$) corresponding to the support set for the $i$-th considered class, and $\mathcal{Q}$ the set of (also preprocessed) query feature vectors.

In the case of inductive few-shot learning, we use a simple Nearest Class Mean classifier (NCM). Predictions are obtained by first computing class barycenters from labeled samples:
\begin{equation}
    \label{eq:barycenter}
    \forall i : \overline{\mathbf{c}_i} = \frac{1}{|\mathcal{S}_{i}|}\sum_{\mathbf{z} \in \mathcal{S}_i}{\mathbf{z}}\;,
\end{equation}
then associating to each query the closest barycenter:
\begin{equation}
    \forall \mathbf{z} \in \mathcal{Q} : C_{ind}(\mathbf{z}, [\overline{\mathbf{c}_1}, \dots, \overline{\mathbf{c}_n}]) = \arg\min_{i}{\left\|\mathbf{z} - \overline{\mathbf{c}_i}\right\|_2}\;.
\end{equation}

In the case of transductive learning, we use a soft K-means algorithm. We compute the following sequence indexed by $t$, where the initial $\overline{\mathbf{c}_i}$ are computed as in Equation (\ref{eq:barycenter}) :
%\begin{equation}
%    \forall i, t :
%    \left\{
%    \begin{array}{lll}\mathbf{c}_i^0 &=& \mathbf{c}_i\;,\\
%    \mathbf{c}_i^{t+1} &=& \textit{weighted mean}( \mathcal{S}_i  \cup \{\mathbf{z} \in %\mathcal{Q},\\
%    & & C_{inductive}(\mathbf{z}, [\mathbf{c}_1^t, \dots, \mathbf{c}_n^t]) = i\})\;.
%    \end{array}
%    \right.
%\end{equation}
%Note that instead of updating class centers with an average that assumes an equal weight for each feature point as in K-means, here we compute a weighted average where weight values are calculated via the L2 distances between data points and class centers, followed by a softmax function adjusted by a value of temperature.

\begin{equation}
    \forall i, t :
    \left\{
    \begin{array}{lll}
    \overline{\mathbf{c}_i}^0 &=& \overline{\mathbf{c}_i}\;,\\
    \overline{\mathbf{c}_i}^{t+1} &=& \sum\limits_{\mathbf{z} \in {\mathcal{S}_i \cup \mathcal{Q}}} \frac{w(\mathbf{z}, \overline{\mathbf{c}_i}^t)}{\sum\limits_{\mathbf{z'} \in {\mathcal{S}_i \cup \mathcal{Q}}} w(\mathbf{z'}, \overline{\mathbf{c}_i}^t)} \mathbf{z}\;,
    
    %\frac{1}{\sum\limits_{\mathbf{z} \in {\mathcal{S}_i^t}} \e^{-\beta\left\|\mathbf{z} - \overline{\mathbf{c}_i}^t \right\|_2^2}} \sum\limits_{\mathbf{z} \in {\mathcal{S}_i^t}} \left(\e^{-\beta\left\|\mathbf{z} - \overline{\mathbf{c}_i}^t \right\|_2^2}\right) \mathbf{z} \;,
    \end{array}
    \right.
\end{equation}
%where $\mathcal{S}_i^t$ is the set of feature vectors pseudo-labeled as $i$:
%\begin{equation}
%    \mathcal{S}_i^t = \mathcal{S}_i \cup \left\{\mathbf{z} \in \mathcal{Q}, C_{ind}(\mathbf{z}, [\overline{\mathbf{c}_1}^{t}, \dots, \overline{\mathbf{c}_n}^{t}]) = i\right\} \;,
%\end{equation}
where $w(\mathbf{z}, \overline{\mathbf{c}_i}^t)$ is a weighting function on $\mathbf{z}$, that gives it a probability of being associated with barycenter $\overline{\mathbf{c}_i}^t$:
\begin{equation}
    w(\mathbf{z}, \overline{\mathbf{c}_i}^t) =
    \left\{
    \begin{array}{lll}
        \frac{\exp\left(-\beta \left\|\mathbf{z} - \overline{\mathbf{c}_i}^t \right\|_2^2\right)}{\sum\limits_{j = 1}^n \exp\left(-\beta \left\|\mathbf{z} - \overline{\mathbf{c}_j}^t \right\|_2^2\right)} & \mathrm{if} & \mathbf{z} \in \mathcal{Q}\;, \\
        1 & \mathrm{if} & \mathbf{z} \in \mathcal{S}_i \;.
    \end{array}
    \right.
\end{equation}

%\begin{equation}
%    \scriptsize
%    \forall i, t :
%    \left\{
%    \begin{array}{lll}
%    \overline{\mathbf{c}_i}^0 &=& \overline{\mathbf{c}_i}\;,\\
%    %\overline{\mathbf{c}_i}^{t+1} &=& \frac{1}{\sum\limits_{\mathbf{z} \in {\mathcal{S}_i^t}} \e^{-\beta\left\|\mathbf{z} - \overline{\mathbf{c}_i}^t \right\|_2^2}} \sum\limits_{\mathbf{z} \in {\mathcal{S}_i^t}} \left(\e^{-\beta\left\|\mathbf{z} - \overline{\mathbf{c}_i}^t \right\|_2^2}\right) \mathbf{z} \;,
%    \overline{\mathbf{c}_i}^{t+1} &=& \frac{1}{\sum\limits_{\mathbf{z} \in {\mathcal{S}_i^t}} \softmax(\mathbf{z}, \overline{\mathbf{c}_i}^t)} \sum\limits_{\mathbf{z} \in {\mathcal{S}_i^t}} \mathbf{z} \cdot \softmax(\mathbf{z}, \overline{\mathbf{c}_i}^t) \;,
%    \end{array}
%    \right.
%\end{equation}
%where
%$softmax(\mathbf{z}, \overline{\mathbf{c}_i}^t) = \frac{\e^{-\beta\left\|\mathbf{z} - \overline{\mathbf{c}_i}^t \right\|_2^2}}{\sum_{j=1}^n \e^{-\beta\left\|\mathbf{z} - \overline{\mathbf{c}_j}^t \right\|_2^2}} $ and
%\begin{equation}
%    \mathcal{S}_i^t = \mathcal{S}_i \cup \left\{\mathbf{z} \in \mathcal{Q}, C_{ind}(\mathbf{z}, [\overline{\mathbf{c}_1}^{t}, \dots, \overline{\mathbf{c}_n}^{t}]) = i\right\} \;.
%\end{equation}

Contrary to the simple K-means algorithm, we use a weighted average where weight values are calculated via a decreasing function of the $L_2$ distance between data points and class barycenters --here, a softmax adjusted by a temperature value $\beta$. In our experiments, we use $\beta = 5$, which led to consistent results across datasets and backbones--. In practice, we use a finite number of steps. By denoting $\mathbf{c}_i^{\infty}$ the resulting vectors, predictions are:
\begin{equation}
    \small
    \forall \mathbf{z} \in \mathcal{Q} : C_{tra}(\mathbf{z}, [\overline{\mathbf{c}_1}^{\infty}, \dots, \overline{\mathbf{c}_n}^\infty]) = \arg\min_{i}{\left\|\mathbf{z} - \overline{\mathbf{c}_i}^{\infty}\right\|_2}\;.
\end{equation}

\subsection{Methods}

In the end, our main method consists of assembling the 5 previously described steps, resulting in the acronym EASY. We have two optional steps, creating the methods Y (without ensemble and augmented samples), ASY (without ensemble) and EY (without augmented samples) for ablation tests.

\section{Results}
\label{sec:results}

\subsection{Ranking on standard benchmarks}
We first report results comparing our method with state of the art using classical settings and datasets. For each method, we specify the number of trainable parameters, the accuracy on 1-shot or 5-shot runs. Experiments always use $q=15$ query samples per class and results are averaged over 10,000 runs. Results are presented in Tables~\ref{tab:results_inductive_miniImageNet}-\ref{tab:results_inductive_tieredImageNet} for the inductive setting and
Tables~\ref{tab:results_transductive_cubfs}-\ref{tab:results_transductive_tieredImageNet} for the transductive setting\footnote{The codes allowing to reproduce our experiments are available at \url{https://github.com/ybendou/easy}.}.

Let us first emphasize that our proposed methodology states a new state-of-the-art performance for MiniImageNet (inductive), TieredImageNet (inductive) and FC100 (both inductive and transductive), while showcasing competitive results on other benchmarks. We believe that, combined with other more elaborate methods, these results could be improved by a fair margin, leading to a new standard of performance for few-shot benchmarks. In the transductive setting, the proposed methodology is less often ranked \#1, but contrary to many alternatives it does not use any prior on class balance in the generated few-shot problems. We provide such experiments in the supplementary material, where we show that the proposed method greatly outperforms existing techniques when considering imbalanced classes.

\subsection{Ablation study}
To better understand the relative contributions of ingredients in the proposed method, we also compare, for each dataset, the performance of various combinations in Table~\ref{tab:ablation} for the inductive setting, and Table~\ref{tab:ablation_trans} for the transductive setting. Interestingly, the full proposed methodology (EASY) is not always the most efficient. We believe that for large datasets such as MiniImageNet and TieredImageNet, the considered ResNet12 backbones contain too few parameters. When reducing this number for ensemble solutions, the drop of performance due to the reduction in size is not compensated by the diversity of the multiple backbones. All things considered, only AS is consistently beneficial to the performance.

%\subsection{Impact of the number of augmented samples}

%\subsection{Impact of the number of backbones}

\section{Conclusion}
In this paper we introduced a very simple method to perform few-shot classification in both inductive and transductive settings. We showed the ability of the method to obtain state of the art results on multiple standardized benchmarks, even beating previous methods by a fair margin in some cases. There is no real new ingredient in this methodology, but we expect it to serve as a baseline for future work.

\input{tables2/inductive/table_inductive_miniImageNet}
\input{tables2/inductive/table_inductive_CUBFS}

\input{tables2/inductive/table_inductive_CIFARFS}
\input{tables2/inductive/table_inductive_FC-100}
\input{tables2/inductive/table_inductive_tiered}

\input{tables2/transductif/miniimagenet}
\input{tables2/transductif/cubfs}

\input{tables2/transductif/cifarfs}

\input{tables2/transductif/fc100}
\input{tables2/transductif/tiered}

\input{tables/ablations/inductive}
\input{tables/ablations/transductive}

\clearpage
\clearpage
%%%%%%%%%%%%%%%%%%%%%%%%%%%%%%%%%%%%%%%%%%%%%%%%%%%%%%%%%%%%
%\bibliographystyle{neurips_template/icml2020}
\bibliographystyle{./IEEEtran}
\bibliography{main}

%%%%%%%%%%%%%%%%%%%%%%%%%%%%%%%%%%%%%%%%%%%%%%%%%

\newpage
~
\newpage

\title{\scalebox{0.93}{EASY -- Ensemble Augmented-Shot Y-shaped Learning:}\\\scalebox{0.78}{State-Of-The-Art Few-Shot Classification with Simple Ingredients}\\\scalebox{0.78}{(Supplementary material)}}

\maketitle

\begin{appendices}
    \input{supplementary_appendix}
\end{appendices}

%%%%%%%%%%%%%%%%%%%%%%%%%%%%%%%%%%%%%%%%%%%%%%%%%

\end{document}

%% file: tikz/sample.tikzstyles
\tikzstyle{blue dot}=[fill={rgb,255: red,127; green,193; blue,255}, draw=none, shape=circle, scale=0.3]
\tikzstyle{red dot}=[fill={rgb,255: red,255; green,127; blue,127}, draw=none, shape=circle, scale=0.3]
\tikzstyle{green dot}=[fill={rgb,255: red,127; green,201; blue,127}, draw=none, shape=circle, scale=0.3]
\tikzstyle{rectangle}=[fill=white, draw=black, shape=rectangle]
\tikzstyle{large box}=[fill=none, draw=black, shape=rectangle, minimum width=0.75cm, minimum height=1cm, rounded corners=2mm]
\tikzstyle{trapezium}=[fill={rgb,255: red,150; green,186; blue,218}, draw=none, shape=trapezium, minimum height=1cm, minimum width=1cm, rotate=270, rounded corners=0.2mm]
\tikzstyle{invisible large box}=[fill=none, draw=none, shape=rectangle, minimum width=0.75cm, minimum height=1cm]
\tikzstyle{dashed circle}=[fill=none, draw=black, shape=circle, minimum size=1.9cm, dashed]
\tikzstyle{star}=[fill=white, draw=black, shape=star, scale=0.3]

\tikzstyle{standard edge}=[draw=black, -, fill={rgb,255: red,215; green,217; blue,218}, fill opacity=0.2]
\tikzstyle{big dash}=[-, thick, dashed, dash pattern=on 4mm off 2mm, fill=none]
\tikzstyle{dashed edge}=[draw=black, -, dashed]
\tikzstyle{arrow}=[->, thick]
\tikzstyle{arrow2}=[{|->}]
\tikzstyle{arrow left}=[<-]
\tikzstyle{arrow left2}=[{<-|}]
\tikzstyle{arrow both sides}=[<->]
\tikzstyle{blue edge}=[-, draw=blue, thick]
\tikzstyle{blue arrow}=[draw=blue, ->, thick]
\tikzstyle{red edge}=[-, draw=red, thick]
\tikzstyle{dash points}=[-, dash pattern=on 0.01mm off 0.1mm, dashed]
\tikzstyle{thin edge}=[-, line width=0.01, fill={rgb,255: red,122; green,190; blue,232}, fill opacity=0.5, thick]
\tikzstyle{gray}=[-, draw={rgb,255: red,191; green,191; blue,191}]
\tikzstyle{edge}=[-, fill=white]

%% file: tables2/inductive/table_inductive_miniImageNet.tex
\begin{table}[!htbp]
    \caption{1-shot and 5-shot accuracy of state-of-the-art methods and proposed solution on \textbf{MiniImageNet} in \textbf{inductive} setting.}
    \centering
    \begin{tabular}{p{0.2cm}p{3.7cm}>{\centering\arraybackslash}p{1.6cm}>{\centering\arraybackslash}p{1.6cm}}
    \toprule
         & Method& 1-shot & 5-shot \\
         \midrule
        \multirow{14}{*}{\rotatebox[origin=c]{90}{$\leq 12M$}
        $\left\{\vphantom{\begin{tabular}{c}~\\~\\~\\~\\~\\~\\~\\~\\~\\~\\~\\~\\~\\~\\\end{tabular}}\right.$}
        & SimpleShot~\cite{wang2019simpleshot}  &$62.85\pm0.20$ & $80.02\pm0.14$\\
        & Baseline++~\cite{Chen2019} & $53.97\pm0.79$ & $75.90\pm0.61$  \\
        & TADAM~\cite{Oreshkin}  & $58.50 \pm 0.30$ & $76.70\pm 0.30$  \\
        & ProtoNet~\cite{Snell2017} &$60.37\pm0.83$ & $78.02\pm0.57$\\
        & R2-D2 (+ens)~\cite{liu2020task} & $64.79 \pm 0.45$ & $81.08\pm0.32$\\
        & FEAT~\cite{Ye2018} & $66.78\quad \quad \quad$ & $82.05\quad  \quad \quad $\\
        & CNL~\cite{zhao2021looking} & $67.96\pm0.98$ & $83.36\pm0.51$\\
        & MERL~\cite{fei2020melr} & $67.40\pm0.43$ & $83.40\pm0.28$\\
        & Deep EMD v2~\cite{zhang2020deepemd} & $68.77\pm0.29$ & $84.13\pm0.53$ \\ 
        & PAL~\cite{ma2021partner}& $69.37\pm0.64$ & $84.40\pm0.44$\\
        & inv-equ~\cite{rizve2021exploring} & $67.28\pm0.80$ & $84.78\pm0.50$\\ 
        & CSEI~\cite{li2021learning}& $68.94\pm0.28 $ & $85.07\pm0.50$\\
        & COSOC~\cite{luo2021rectifying} & $69.28\pm0.49$ & $85.16\pm0.42$\\
        & EASY 2$\times$ResNet12$\left(\frac{1}{\sqrt{2}}\right)$ \scriptsize{(ours)}  & $\mathbf{70.63 \pm 0.20}$ & $\mathbf{86.28 \pm 0.12}$\\
        \midrule
        \multirow{3}{*}{\rotatebox[origin=c]{90}{$36M$}
        $\left\{\vphantom{\begin{tabular}{c}~\\~\\~\\\end{tabular}}\right.$}
        & S2M2R~\cite{mangla2020charting} &$64.93\pm0.18$ & $83.18\pm0.11$\\
        & LR + DC~\cite{yang2021free}&$68.55\pm0.55$ & $82.88\pm0.42$\\
        & EASY 3$\times$ResNet12 \scriptsize{(ours)} & $\mathbf{71.75 \pm0.19}$ & $\mathbf{87.15 \pm0.12}$\\
        \bottomrule
    \end{tabular}
    \label{tab:results_inductive_miniImageNet}
\end{table}  

% TADAM: Oreshkin, MAML:Finn2017

%% file: tables2/inductive/table_inductive_CUBFS.tex
\begin{table}[!htbp]
    \caption{1-shot and 5-shot accuracy of state-of-the-art methods and proposed solution on \textbf{CUB-FS} in \textbf{inductive} setting.}
    \centering
    \begin{tabular}{p{0.2cm}p{3.7cm}>{\centering\arraybackslash}p{1.6cm}>{\centering\arraybackslash}p{1.6cm}}
    \toprule
         & Method& 1-shot & 5-shot \\
         \midrule
        \multirow{5}{*}{\rotatebox[origin=c]{90}{$\leq 12M$}
        $\left\{\vphantom{\begin{tabular}{c}~\\~\\~\\~\\~\\\end{tabular}}\right.$}
        & FEAT~\cite{Ye2018} & $68.87 \pm 0.22$ & $82.90\pm0.10$\\ % FEAT %113K parameters for conv4
         & LaplacianShot~\cite{MasudZiko2020} & $\mathbf{80.96\quad\quad\quad}$ & $88.68\quad\quad\quad$ \\ % Protonet

         & ProtoNet~\cite{Snell2017}  & $66.09\pm0.92$ & $82.50\pm 0.58$ \\ % Protonet
          & DeepEMD v2~\cite{zhang2020deepemd}& $79.27\pm 0.29$ & $89.80\pm 0.51$ \\ % DeepEMD
          & EASY 4$\times$ResNet12$\left(\frac{1}{2}\right)$ \scriptsize{(ours)}  & $77.97 \pm 0.20$  & $\mathbf{91.59 \pm 0.10}$\\
       \midrule
        \multirow{2}{*}{\rotatebox[origin=c]{90}{$36M$}
        $\left\{\vphantom{\begin{tabular}{c}~\\~\\\end{tabular}}\right.$}
        & S2M2R~\cite{mangla2020charting}  &$\mathbf{80.68\pm0.81}$ & $90.85\pm0.44$\\ %S2M2R
        & EASY 3$\times$ResNet12 \scriptsize{(ours)} & $78.56 \pm 0.19$ & $\mathbf{91.93 \pm 0.10}$\\
         
    \bottomrule
    \end{tabular}
    \label{tab:results_inductive_CUBFS}
\end{table}

%% file: tables2/inductive/table_inductive_CIFARFS.tex
\begin{table}[!htbp]
    \caption{1-shot and 5-shot accuracy of state-of-the-art methods and proposed solution on \textbf{CIFAR-FS} in \textbf{inductive} setting.}
    \centering
    \begin{tabular}{p{0.2cm}p{3.7cm}>{\centering\arraybackslash}p{1.6cm}>{\centering\arraybackslash}p{1.6cm}}
    \toprule
         & Method& 1-shot & 5-shot \\
         \midrule
        \multirow{4}{*}{\rotatebox[origin=c]{90}{$\leq 12M$}
        $\left\{\vphantom{\begin{tabular}{c}~\\~\\~\\~\\\end{tabular}}\right.$}
        &S2M2R~\cite{mangla2020charting} & $63.66\pm 0.17$ & $76.07 \pm 0.19$ \\ % S2M2R
         
          &R2-D2 (+ens)~\cite{liu2020task}& $76.51\pm 0.47$ & $87.63 \pm 0.34$ \\ % R2-D2 (+ens) (+val)
         &invariance-equivariance~\cite{rizve2021exploring} & $\mathbf{77.87\pm0.85}$ & $\mathbf{89.74 \pm 0.57}$\\ %equivariance invariance
          &EASY 2$\times$ResNet12$\left(\frac{1}{\sqrt{2}}\right)$ \scriptsize{(ours)}  & $75.24 \pm 0.20$ & $88.38 \pm 0.14$\\
       \midrule
        \multirow{2}{*}{\rotatebox[origin=c]{90}{$36M$}
        $\left\{\vphantom{\begin{tabular}{c}~\\~\\\end{tabular}}\right.$}
        &S2M2R~\cite{mangla2020charting} &$74.81\pm0.19$ & $87.47\pm0.13$\\ %S2M2R
         &EASY 3$\times$ResNet12 \scriptsize{(ours)} & $\mathbf{76.20\pm 0.20}$ & $\mathbf{89.00 \pm 0.14}$\\
         
    \bottomrule
    \end{tabular}
    \label{tab:results_inductive_CIFARFS}
\end{table}

 % xu2020attentional 75.4 ± 0.2 86.8 ± 0.2
 
 % tian2020rethinking 73.9 ± 0.8 86.9 ± 0.5

%% file: tables2/inductive/table_inductive_FC-100.tex
\begin{table}[!htbp]
    \caption{1-shot and 5-shot accuracy of state-of-the-art methods and proposed solution on \textbf{FC-100} in \textbf{inductive} setting.}
    \centering
    \begin{tabular}{p{0.2cm}p{3.7cm}>{\centering\arraybackslash}p{1.6cm}>{\centering\arraybackslash}p{1.6cm}}
    \toprule
         & Method& 1-shot & 5-shot \\
         \midrule
        \multirow{6}{*}{\rotatebox[origin=c]{90}{$\leq 12M$}
        $\left\{\vphantom{\begin{tabular}{c}~\\~\\~\\~\\~\\~\\\end{tabular}}\right.$}
        &DeepEMD v2~\cite{zhang2020deepemd} &$46.60\pm0.26  $&$ 63.22 \pm 0.71$ \\ 
        &TADAM~\cite{Oreshkin}  & $40.10 \pm 0.40$ & $56.10\pm 0.40$ \\
        &ProtoNet~\cite{Snell2017}  & $41.54 \pm 0.76$ & $57.08\pm 0.76$ \\ %ProtoNet
        &invariance-equivariance~\cite{rizve2021exploring}  & $47.76  \pm 0.77$ & $\mathbf{65.30 \pm 0.76 }$\\ %equivariance invariance
        &R2-D2 (+ens)~\cite{liu2020task} &  $44.75\pm 0.43$ & $59.94\pm0.41 $ \\ 
        &EASY 2$\times$ResNet12$\left(\frac{1}{\sqrt{2}}\right)$ \scriptsize{(ours)} & $\mathbf{47.94 \pm 0.19}$ & $64.14 \pm 0.19$\\
        \midrule
        \multirow{2}{*}{\rotatebox[origin=c]{90}{$36M$}
        $\left\{\vphantom{\begin{tabular}{c}~\\~\\\end{tabular}}\right.$}
        & & & \\
        & EASY 3$\times$ResNet12 \scriptsize{(ours)} & $\mathbf{48.07 \pm 0.19}$ & $64.74 \pm 0.19$\\
        & & & \\
    \bottomrule
    \end{tabular}
    \label{tab:results_inductive_FC100}
\end{table}

%% file: tables2/inductive/table_inductive_tiered.tex
\begin{table}[!htbp]
    \caption{1-shot and 5-shot accuracy of state-of-the-art methods and proposed solution on \textbf{TieredImageNet} in \textbf{inductive} setting.}
    \centering
    \begin{tabular}{p{0.2cm}p{3.7cm}>{\centering\arraybackslash}p{1.6cm}>{\centering\arraybackslash}p{1.6cm}}
    \toprule
         & Method& 1-shot & 5-shot \\
         \midrule
        \multirow{11}{*}{\rotatebox[origin=c]{90}{$\leq 12M$}
        $\left\{\vphantom{\begin{tabular}{c}~\\~\\~\\~\\~\\~\\~\\~\\~\\~\\~\\\end{tabular}}\right.$}
        &SimpleShot~\cite{wang2019simpleshot} & $69.09\pm0.22$ & $84.58\pm0.16$ \\ %simpleshots
        
        &ProtoNet~\cite{Snell2017} &$65.65\pm0.92$ & $83.40\pm0.65$\\ %ProtoNet
        
        &FEAT~\cite{Ye2018} &$70.80\pm 0.23 $& $84.79\pm 0.16$ \\ %FEAT
        &PAL~\cite{ma2021partner} &$72.25\pm0.72$ & $86.95\pm 0.47$\\ %PAL
        &DeepEMD v2~\cite{zhang2020deepemd} &$74.29\pm 0.32$ & $86.98\pm0.60$ \\ 
        &MERL~\cite{fei2020melr} &$72.14\pm0.51 $ & $87.01\pm 0.35$ \\ %MERL
        
        &COSOC~\cite{luo2021rectifying} & $73.57\pm0.43$ & $87.57\pm0.10$\\ %COSOC
        &CNL~\cite{zhao2021looking} & $73.42 \pm 0.95$  &$87.72\pm 0.75$ \\ %CNL

        &invariance-equivariance~\cite{rizve2021exploring} &$72.21\pm 0.90$&$87.08\pm0.58$\\   
        &CSEI~\cite{li2021learning} &$73.76\pm0.32$ & $87.83\pm0.59$ \\ %CSEI
        &ASY ResNet12 \scriptsize{(ours)} & $\mathbf{74.31\pm 0.22}$ & $\mathbf{87.86 \pm 0.15}$\\

         \midrule
        \multirow{2}{*}{\rotatebox[origin=c]{90}{$36M$}
        $\left\{\vphantom{\begin{tabular}{c}~\\~\\\end{tabular}}\right.$}
        &S2M2R~\cite{mangla2020charting} &$73.71\pm 0.22$ & $\mathbf{88.52\pm0.14}$\\ %S2M2R
        &EASY 3$\times$ResNet12 \scriptsize{(ours)} & $\mathbf{74.71 \pm0.22}$ & $88.33 \pm 0.14$\\

    \bottomrule
    \end{tabular}
    \label{tab:results_inductive_tieredImageNet}
\end{table}

%% file: tables2/transductif/miniimagenet.tex
\begin{table}[!htbp]
    \caption{1-shot and 5-shot accuracy of state-of-the-art methods and proposed solution on \textbf{MiniImageNet} in \textbf{transductive} setting.}
    \centering
    \begin{tabular}{p{0.2cm}p{3.7cm}>{\centering\arraybackslash}p{1.6cm}>{\centering\arraybackslash}p{1.6cm}}
    \toprule
         & Method& 1-shot & 5-shot \\
         \midrule
        \multirow{10}{*}{\rotatebox[origin=c]{90}{$\leq 12M$}
        $\left\{\vphantom{\begin{tabular}{c}~\\~\\~\\~\\~\\~\\~\\~\\~\\~\\\end{tabular}}\right.$}
        & TIM-GD~\cite{boudiaf2020transductive} &$73.90\quad\quad\quad$ & $85.00\quad\quad\quad$\\ 
        & ODC~\cite{Qi2021} &$77.20\pm0.36$ & $87.11\pm0.42$\\ 
        & PEM$_n$E-BMS$^*$ ~\cite{hu2021squeezing}  &$80.56\pm0.27$ & $87.98\pm0.14$\\
        & SSR~\cite{Shen} & $ 68.10\pm0.60$  & $76.90\pm0.40 $ \\
        & iLPC~\cite{Lazarou} & $69.79\pm 0.99$ &$79.82\pm0.55$ \\
        
        & EPNet~\cite{rodriguez2020embedding}& $66.50\pm 0.89$ &$81.60\pm0.60$ \\
        & DPGN~\cite{Yang_2020_CVPR} & $67.77\pm0.32$  & $84.60\pm0.43 $ \\
        & ECKPN~\cite{Chen2021} & $70.48\pm0.38$  & $85.42\pm0.46 $ \\
        & Rot+KD+POODLE~\cite{Le2021} & $77.56\quad\quad\quad$ &$85.81\quad\quad\quad$ \\
        %& EASY 2$\times$ResNet12($\frac{1}{\sqrt{2}}$)^*  & $78.46 \pm 0.24$& $87.95 \pm 0.12$\\
        & EASY 2$\times$ResNet12$\left(\frac{1}{\sqrt{2}}\right)$ \scriptsize{(ours)}  & $\mathbf{82.31 \pm 0.24}$& $\mathbf{88.57 \pm 0.12}$\\

         \midrule
         \multirow{11}{*}{\rotatebox[origin=c]{90}{$36M$}
        $\left\{\vphantom{\begin{tabular}{c}~\\~\\~\\~\\~\\~\\~\\~\\~\\~\\~\\\end{tabular}}\right.$}
        & SSR~\cite{Shen} & $ 72.40\pm0.60$  & $80.20\pm0.40 $ \\
        & fine-tuning(train+val)~\cite{Dhillon2019} &$68.11\pm0.69$ & $80.36\pm0.50$\\ 
        & SIB+E$^3$BM~\cite{Liu2020} & $71.40\quad\quad\quad$ & $81.20\quad\quad\quad$\\ 
        & LR+DC~\cite{yang2021free} & $ 68.57\pm0.55$  & $82.88\pm0.42 $ \\
        & EPNet~\cite{rodriguez2020embedding} & $70.74\pm 0.85$ &$84.34\pm0.53$ \\
        & TIM-GD~\cite{boudiaf2020transductive}  &$77.80\quad\quad\quad$ & $87.40\quad\quad\quad$\\ 
        & PT+MAP~\cite{Hu2021}  & $82.92\pm 0.26$ &$88.82\pm0.13$ \\
        & iLPC~\cite{Lazarou}   & $83.05\pm 0.79$ &$88.82\pm0.42$ \\
        & ODC~\cite{Qi2021}    &$80.64\pm0.34$ & $89.39\pm0.39$\\ 
        
        & PEM$_n$E-BMS$^*$~\cite{hu2021squeezing}  &$83.35\pm0.25$ & $\mathbf{89.53\pm0.13}$\\ 
        %& EASY 3$\times$ResNet12^{*}  & $79.44 \pm 0.24$& $88.43 \pm 0.12$\\
        & EASY 3$\times$ResNet12 \scriptsize{(ours)} &$\mathbf{84.04 \pm 0.23}$& $89.14 \pm 0.11$\\ 
    \bottomrule
    \end{tabular}
    \label{tab:results_transductive_miniImageNet}
\end{table}

%% file: tables2/transductif/cubfs.tex
\begin{table}[!htbp]
    \caption{1-shot and 5-shot accuracy of state-of-the-art methods and proposed solution on \textbf{CUB-FS} in \textbf{transductive} setting.}
    \centering
    \begin{tabular}{p{0.2cm}p{3.7cm}>{\centering\arraybackslash}p{1.6cm}>{\centering\arraybackslash}p{1.6cm}}
    \toprule
         & Method& 1-shot & 5-shot \\
         \midrule
        \multirow{7}{*}{\rotatebox[origin=c]{90}{$\leq 12M$}
        $\left\{\vphantom{\begin{tabular}{c}~\\~\\~\\~\\~\\~\\~\\\end{tabular}}\right.$}
        & TIM-GD~\cite{boudiaf2020transductive} &$82.20\quad\quad\quad$ & $90.80\quad\quad\quad$\\ 
        & ODC~\cite{Qi2021} &$85.87\quad\quad\quad$ & $94.97\quad\quad\quad$\\
        
        & DPGN~\cite{Yang_2020_CVPR} & $75.71\pm0.47$  & $91.48\pm0.33 $ \\
        & ECKPN~\cite{Chen2021} & $77.43\pm0.54$  & $92.21\pm0.41 $ \\
        & iLPC~\cite{Lazarou} & $89.00\pm 0.70$ &$92.74\pm0.35$ \\
        & Rot+KD+POODLE~\cite{Le2021} & $89.93\quad\quad\quad$ &$\mathbf{93.78\quad\quad\quad}$ \\
        %& EASY 4$\times$ResNet12($\frac{1}{2}$)^* & $87.53 \pm 0.21$ & $92.92 \pm 0.10$\\
        & EASY 4$\times$ResNet12$\left(\frac{1}{2}\right)$ \scriptsize{(ours)} & $\mathbf{90.50 \pm 0.19}$ & $93.50 \pm 0.09$\\

        \midrule
        \multirow{4}{*}{\rotatebox[origin=c]{90}{$36M$}
        $\left\{\vphantom{\begin{tabular}{c}~\\~\\~\\~\\\end{tabular}}\right.$}
        & LR+DC~\cite{yang2021free} & $ 79.56\pm0.87$  & $90.67\pm0.35 $ \\
        & PT+MAP~\cite{Hu2021} & $\mathbf{91.55\pm 0.19}$ &$93.99\pm0.10$ \\
        & iLPC~\cite{Lazarou} & $91.03\pm 0.63$ &$\mathbf{94.11\pm0.30}$ \\
        %& EASY 3$\times$ResNet12^* & $87.67 \pm 0.21$ & $93.30 \pm 0.09$\\
        & EASY 3$\times$ResNet12 \scriptsize{(ours)} & $90.56 \pm 0.19$ & $93.79 \pm 0.10$\\

    \bottomrule
    \end{tabular}
    \label{tab:results_transductive_cubfs}
\end{table}

%% file: tables2/transductif/cifarfs.tex
\begin{table}[!htbp]
    \caption{1-shot and 5-shot accuracy of state-of-the-art methods and proposed solution on \textbf{CIFAR-FS} in \textbf{transductive} setting.}
    \centering
    \begin{tabular}{p{0.2cm}p{3.7cm}>{\centering\arraybackslash}p{1.6cm}>{\centering\arraybackslash}p{1.6cm}}
    \toprule
         & Method& 1-shot & 5-shot \\
         \midrule
        \multirow{5}{*}{\rotatebox[origin=c]{90}{$\leq 12M$}
        $\left\{\vphantom{\begin{tabular}{c}~\\~\\~\\~\\~\\\end{tabular}}\right.$}
        &SSR~\cite{Shen} & $76.80\pm0.60$  & $83.70\pm0.40 $ \\
        &iLPC~\cite{Lazarou}& $77.14\pm 0.95$ &$85.23\pm0.55$ \\
        &DPGN~\cite{Yang_2020_CVPR} & $77.90\pm0.50$  & $90.02\pm0.40 $ \\
        &ECKPN~\cite{Chen2021} & $79.20\pm0.40$  & $\mathbf{91.00\pm0.50} $ \\
        %&EASY 2$\times$ResNet12($\frac{1}{\sqrt{2}}$)^* & $\mathbf{83.56 \pm 0.23}$ & $89.64 \pm 0.15$\\
        &EASY 2$\times$ResNet12$\left(\frac{1}{\sqrt{2}}\right)$ \scriptsize{(ours)} & $\mathbf{86.99 \pm 0.21}$ & $90.20 \pm 0.15$\\
         \midrule
        \multirow{5}{*}{\rotatebox[origin=c]{90}{$36M$}
        $\left\{\vphantom{\begin{tabular}{c}~\\~\\~\\~\\~\\\end{tabular}}\right.$}
        &SSR~\cite{Shen} & $81.60\pm0.60$  & $86.00\pm0.40 $ \\
        &fine-tuning (train+val)~\cite{Dhillon2019} &$78.36\pm0.70$ & $87.54\pm0.49$\\
        &iLPC~\cite{Lazarou}& $86.51\pm 0.75$ &$90.60\pm0.48$ \\
        &PT+MAP~\cite{Hu2021}& $\mathbf{87.69\pm 0.23}$ &$\mathbf{90.68\pm0.15}$ \\
        %&EASY 3$\times$ResNet12^* & $84.60\pm 0.22$ & $90.12 \pm 0.15$\\
        &EASY 3$\times$ResNet12 \scriptsize{(ours)} & $87.16\pm 0.21$ & $90.47 \pm 0.15$\\
        
    \bottomrule
    \end{tabular}
    \label{tab:results_transductive_FC100}
\end{table}

%% file: tables2/transductif/fc100.tex
\begin{table}[!htbp]
    \caption{1-shot and 5-shot accuracy of state-of-the-art methodsand proposed solution on \textbf{FC-100} in \textbf{transductive} setting.}
    \centering
    \begin{tabular}{p{0.2cm}p{3.7cm}>{\centering\arraybackslash}p{1.6cm}>{\centering\arraybackslash}p{1.6cm}}
    \toprule
         & Method& 1-shot & 5-shot \\
         \midrule
        \multirow{3}{*}{\rotatebox[origin=c]{90}{$\leq 12M$}
        $\left\{\vphantom{\begin{tabular}{c}~\\~\\~\\\end{tabular}}\right.$}
        &&&\\
        
         %\begin{turn}{90}\hspace{-0.2cm} \scriptsize{$12M$}\end{turn}&EASY 2$\times$ResNet12($\frac{1}{\sqrt{2}}$)^* & $\mathbf{52.19 \pm 0.25}$ & $\mathbf{65.74\pm 0.20}$\\
         &EASY 2$\times$ResNet12$\left(\frac{1}{\sqrt{2}}\right)$ \scriptsize{(ours)} & $\mathbf{54.47 \pm 0.24}$ & $\mathbf{65.82\pm 0.19}$\\
         &&&\\
         \midrule
        \multirow{5}{*}{\rotatebox[origin=c]{90}{$36M$}
        $\left\{\vphantom{\begin{tabular}{c}~\\~\\~\\~\\~\\\end{tabular}}\right.$}
        &SIB+E$^3$BM~\cite{Liu2020} & $46.00\quad\quad\quad$ & $57.10\quad\quad\quad$\\
        &fine-tuning (train)~\cite{Dhillon2019}  &$43.16\pm0.59$ & $57.57\pm0.55$\\
        &ODC~\cite{Qi2021} &$47.18\pm0.30$ & $59.21\pm0.56$\\ 
        &fine-tuning (train+val)~\cite{Dhillon2019} &$50.44\pm0.68$ & $65.74\pm0.60$\\

        %&EASY 3$\times$ResNet12^* & $\mathbf{52.18 \pm 0.25}$& $\mathbf{66.35 \pm 0.20}$\\
        &EASY 3$\times$ResNet12 \scriptsize{(ours)} & $\mathbf{54.13 \pm 0.24}$& $\mathbf{66.86 \pm 0.19}$\\

    \bottomrule
    \end{tabular}
    \label{tab:results_transductive_FC100}
\end{table}

%% file: tables2/transductif/tiered.tex
\begin{table}[!htbp]
    \caption{1-shot and 5-shot accuracy of state-of-the-art methods and proposed solution on \textbf{TieredImageNet} in \textbf{transductive} setting.}
    \centering
    \begin{tabular}{p{0.2cm}p{3.7cm}>{\centering\arraybackslash}p{1.6cm}>{\centering\arraybackslash}p{1.6cm}}
    \toprule
         & Method& 1-shot & 5-shot \\
         \midrule
        \multirow{10}{*}{\rotatebox[origin=c]{90}{$\leq 12M$}
        $\left\{\vphantom{\begin{tabular}{c}~\\~\\~\\~\\~\\~\\~\\~\\~\\~\\\end{tabular}}\right.$}
        & PT+MAP~\cite{Hu2021} &$85.67\pm 0.26$ &$90.45\pm0.14$ \\
        
        & TIM-GD~\cite{boudiaf2020transductive} &$79.90\quad\quad\quad$ & $88.50\quad\quad\quad$\\ 
        & ODC~\cite{Qi2021}  &$83.73\pm0.36$ & $\mathbf{90.46\pm0.46}$\\ 
        
        & SSR~\cite{Shen} & $81.20\pm0.60$  & $85.70\pm0.40 $ \\
        & Rot+KD+POODLE~\cite{Le2021} & $79.67\quad\quad\quad$ &$86.96\quad\quad\quad$ \\
        & DPGN~\cite{Yang_2020_CVPR} & $72.45\pm0.51$  & $87.24\pm0.39 $ \\
        & EPNet~\cite{rodriguez2020embedding} & $76.53\pm 0.87$ &$87.32\pm0.64$ \\
        & ECKPN~\cite{Chen2021} & $73.59\pm0.45$  & $88.13\pm0.28 $ \\
        & iLPC~\cite{Lazarou} & $83.49\pm 0.88$ &$89.48\pm0.47$ \\
        %& ASY ResNet12^* & $80.75\pm 0.27$ & $88.86 \pm 0.14$\\
        & ASY ResNet12 \scriptsize{(ours)} & $\mathbf{83.98\pm 0.24}$ & $89.26 \pm 0.14$\\
        
        \midrule
        \multirow{10}{*}{\rotatebox[origin=c]{90}{$36M$}
        $\left\{\vphantom{\begin{tabular}{c}~\\~\\~\\~\\~\\~\\~\\~\\~\\~\\\end{tabular}}\right.$}
        & SIB+E$^3$BM~\cite{Liu2020} & $75.60\quad\quad\quad$ & $84.30\quad\quad\quad$\\
        & SSR~\cite{Shen} & $79.50\pm0.60$  & $84.80\pm0.40 $ \\
        & fine-tuning (train+val)~\cite{Dhillon2019}  &$72.87\pm0.71$ & $86.15\pm0.50$\\ 
        & TIM-GD~\cite{boudiaf2020transductive}  &$82.10\quad\quad\quad$ & $89.80\quad\quad\quad$\\ 
        & LR+DC~\cite{yang2021free} & $ 78.19\pm0.25$  & $89.90\pm0.41 $ \\ %LR+DC (trasductif)
        & EPNet~\cite{rodriguez2020embedding} & $78.50\pm 0.91$ &$88.36\pm0.57$ \\
        & ODC~\cite{Qi2021}  &$85.22\pm0.34$ & $91.35\pm0.42$\\ 
        & iLPC~\cite{Lazarou} & $\mathbf{88.50\pm 0.75}$ &$\mathbf{92.46\pm0.42}$ \\
        & PEM$_n$E-BMS$^*$~\cite{hu2021squeezing} &$86.07\pm0.25$ & $91.09\pm0.14$\\
        %& EASY 3$\times$ResNet12^*  & $81.01 \pm 0.27$ & $89.29 \pm 0.14$\\
        & EASY 3$\times$ResNet12 \scriptsize{(ours)}  & $84.29 \pm 0.24$ & $89.76 \pm 0.14$\\

    \bottomrule
    \end{tabular}
    \label{tab:results_transductive_tieredImageNet}
\end{table}

%% file: tables/ablations/inductive.tex
\begin{table}[ht]
    \caption{Ablation study of the steps of proposed solution in \textbf{inductive} setting, for a fixed number of trainable parameters in the considered backbones. When using ensembles, we use 2$\times$ResNet12$\left(\frac{1}{\sqrt{2}}\right)$ instead of a single ResNet12.}
    \label{tab:ablation}
    \centering
    \begin{tabular}{ccccc}
    \toprule
    Dataset & E & AS & 1-shot & 5-shot\\
    \midrule
    \multirow{4}*{MiniImageNet} 
    & & & $68.43 \pm 0.19$ & $83.78 \pm 0.13$\\
    & & \checkmark & $\mathbf{70.84 \pm 0.19}$ & $85.70\pm 0.13$\\
    & \checkmark & & $68.69\pm 0.20$& $84.84 \pm 0.13$\\
    & \checkmark & \checkmark & $70.63\pm 0.20$& $\mathbf{86.28 \pm 0.12}$\\
    \midrule
    \multirow{4}*{CUB-FS} 
    & & & $74.13 \pm 0.20$ & $89.08 \pm 0.11$ \\
    & & \checkmark & $77.40 \pm 0.20$ & $\mathbf{91.15 \pm 0.10}$\\
    & \checkmark & & $75.01 \pm 0.20$ & $89.38 \pm 0.11$\\
    & \checkmark & \checkmark & $\mathbf{77.59 \pm 0.20}$ & $91.07 \pm 0.11$\\
    \midrule
    \multirow{4}*{CIFAR-FS} 
    & & & $73.38 \pm 0.21$ & $87.42\pm 0.15$\\
    & & \checkmark & $74.26 \pm 0.21$ & $88.16 \pm 0.15$\\
    & \checkmark & & $74.36 \pm 0.21$& $87.82\pm 0.15$\\
    & \checkmark & \checkmark & $\mathbf{75.24 \pm 0.20}$& $\mathbf{88.38 \pm 0.14}$\\
    \midrule
    \multirow{4}*{FC-100} 
    & & & $45.68 \pm 0.19$ & $62.78 \pm 0.19$\\
    & & \checkmark & $46.43\pm 0.19$ & $64.16 \pm 0.19$\\
    & \checkmark & & $47.52 \pm 0.19$ & $63.92 \pm 0.19$\\
    & \checkmark & \checkmark & $\mathbf{47.94 \pm 0.20}$ & $\mathbf{64.14 \pm 0.19}$\\
    \midrule
    \multirow{4}*{TieredImageNet} 
    & & & $72.52 \pm 0.22$ & $86.79 \pm 0.15$\\
    & & \checkmark & $ \mathbf{74.17 \pm 0.22}$ & $\mathbf{87.81\pm 0.14}$\\
    & \checkmark & & $72.14 \pm 0.22$ & $86.66\pm 0.15$\\
    & \checkmark & \checkmark & $73.36 \pm 0.22$&  $87.37\pm 0.15$\\

    \bottomrule
    \end{tabular}
\end{table}

%% file: tables/ablations/transductive.tex
\begin{table}[!htbp]
    \caption{Ablation study of the steps of propsoed solution in \textbf{transductive} setting, for a fixed number of trainable parameters in the considered backbones. When using ensembles, we use 2$\times$ResNet12$\left(\frac{1}{\sqrt{2}}\right)$ instead of a single ResNet12.}
    \label{tab:ablation_trans}
    \centering
    \begin{tabular}{ccccc}
    \toprule
    Dataset & E & AS & 1-shot & 5-shot\\
    \midrule
    \multirow{4}*{MiniImageNet} 
    & & & $80.42 \pm 0.23$ & $86.72\pm 0.13$ \\
    & & \checkmark & $\mathbf{83.02 \pm 0.23}$ & $88.36 \pm 0.12$\\
    & \checkmark & & $80.27 \pm 0.23$ & $87.45 \pm 0.12$\\
    & \checkmark & \checkmark & $82.31\pm 0.24$ & $\mathbf{88.57\pm 0.12}$\\
    \midrule
    \multirow{4}*{CUB-FS} 
    & & & $86.93 \pm 0.21$ & $91.53\pm 0.11$ \\
    & & \checkmark & $89.80\pm 0.20$ & $93.12\pm 0.10$\\
    & \checkmark & & $87.28 \pm 0.21$& $91.89\pm 0.10$\\
    & \checkmark & \checkmark & $\mathbf{90.05\pm 0.19}$&$\mathbf{93.17\pm 0.10}$ \\
    \midrule
    \multirow{4}*{CIFAR-FS} 
    & & & $84.18 \pm 0.23$ & $89.56 \pm 0.15$\\
    & & \checkmark & $85.55\pm 0.23$ & $90.07\pm 0.15$\\
    & \checkmark & & $84.89 \pm 0.22$ & $89.60 \pm 0.15$ \\
    & \checkmark & \checkmark & $\mathbf{86.99\pm 0.21}$ & $\mathbf{90.20 \pm 0.15}$\\
    \midrule
    \multirow{4}*{FC-100} 
    & & & $51.74\pm 0.23$ & $65.39\pm 0.19$\\
    & & \checkmark & $52.93\pm 0.23$ & $\mathbf{66.51\pm 0.19}$\\
    & \checkmark & & $53.39 \pm 0.23$ & $65.71\pm 0.19$\\
    & \checkmark & \checkmark & $\mathbf{54.47 \pm 0.24}$& $65.82\pm 0.19$\\
    \midrule
    \multirow{4}*{TieredImageNet} 
    & & & $82.32 \pm 0.24$ & $88.45\pm 0.15$\\
    & & \checkmark & $\mathbf{83.98 \pm 0.24}$ & $\mathbf{89.26 \pm 0.14}$\\
    & \checkmark & & $81.48\pm 0.25$ & $88.40 \pm 0.15$ \\
    & \checkmark & \checkmark & $83.20 \pm 0.25$ & $88.92\pm 0.14$\\
    \bottomrule
    \end{tabular}
\end{table}

%% file: supplementary_appendix.tex
\section{Transductive tests with imbalanced settings}

Following the methodology recently proposed in~\cite{veilleux2021realistic}, we also report performance in transductive setting when the number of query vectors is varying for each class and is unknown. Results are presented in Tables~\ref{tab:results_imbalanced_transductive_miniImageNet}-\ref{tab:results_imbalanced_transductive_cubfs}. We note that the proposed methodology is able to outperform existing ones by a fair margin.

\begin{table}[!htbp]
    \caption{1-shot and 5-shot accuracy of state-of-the-art methods and proposed solution on \textbf{MiniImageNet} in \textbf{imbalanced transductive} setting.}
    \centering
    \small
    \begin{tabular}{p{0.2cm}p{3.7cm}>{\centering\arraybackslash}p{1.6cm}>{\centering\arraybackslash}p{1.6cm}}
    \toprule
         & Method& 1-shot & 5-shot \\
         \midrule
        \multirow{6}{*}{\rotatebox[origin=c]{90}{$\leq 12M$}
        $\left\{\vphantom{\begin{tabular}{c}~\\~\\~\\~\\~\\~\\\end{tabular}}\right.$}
        &MAML~\cite{finn2017model} &$47.6$ & $64.5$\\ 
        &LR+ICI~\cite{wang2020instance}  &$58.7$ & $73.5$\\ 
        &PT+MAP~\cite{hu2020leveraging}  &$60.1$ & $67.1$\\ 
        &LaplacianShot~\cite{MasudZiko2020}  &$65.4$ & $81.6$\\ 
        &TIM~\cite{boudiaf2020transductive}  &$67.3$ & $79.8$\\ 
        &$\alpha$-TIM~\cite{veilleux2021realistic}  &$67.4$ & $82.5$\\ 
        \midrule
        \multirow{6}{*}{\rotatebox[origin=c]{90}{$36M$}
        $\left\{\vphantom{\begin{tabular}{c}~\\~\\~\\~\\~\\~\\\end{tabular}}\right.$}
        &PT+MAP~\cite{hu2020leveraging}  & $60.6$ &$66.8$ \\
        &SIB~\cite{hu2020empirical}  & $64.7$ &$72.5$ \\
        &LaplacianShot~\cite{MasudZiko2020}   & $68.1$ &$83.2$ \\
        &TIM~\cite{boudiaf2020transductive}  &$69.8$ & $81.6$\\ 
        &$\alpha$-TIM~\cite{veilleux2021realistic} &$69.8$ & $84.8$\\ 
        &EASY 3$\times$ResNet12 \scriptsize{(ours)} & $\mathbf{76.04}$ & $\mathbf{87.23}$\\
    \bottomrule
    \end{tabular}
    \label{tab:results_imbalanced_transductive_miniImageNet}
\end{table}

\begin{table}[!htbp]
    \caption{1-shot and 5-shot accuracy of state-of-the-art methods and proposed solution on \textbf{TieredImageNet} in \textbf{imbalanced transductive} setting.}
    \centering
    \small
    \begin{tabular}{p{0.2cm}p{3.7cm}>{\centering\arraybackslash}p{1.6cm}>{\centering\arraybackslash}p{1.6cm}}
    \toprule
        & Method & 1-shot & 5-shot \\
        \midrule
         \multirow{4}{*}{\rotatebox[origin=c]{90}{$\leq 12M$}
        $\left\{\vphantom{\begin{tabular}{c}~\\~\\~\\~\\~\\~\\\end{tabular}}\right.$}
        &Entropy-min~\cite{dhillon2019baseline}   &$61.2$ & $75.5$\\ 
        &PT+MAP~\cite{hu2020leveraging}  &$64.1$ & $70.0$\\ 
        &LaplacianShot~\cite{MasudZiko2020} &$72.3$ & $85.7$\\ 
        &TIM~\cite{boudiaf2020transductive}  &$74.1$ & $84.1$\\ 
        &LR+ICI~\cite{wang2020instance}  &$74.6$ & $85.1$\\ 
        &$\alpha$-TIM~\cite{veilleux2021realistic} &$74.4$ & $86.6$\\ 
         \midrule
          \multirow{5}{*}{\rotatebox[origin=c]{90}{$36M$}
        $\left\{\vphantom{\begin{tabular}{c}~\\~\\~\\~\\~\\~\\\end{tabular}}\right.$}
        &Entropy-min~\cite{dhillon2019baseline} &$62.9$ & $77.3$\\
        &PT+MAP~\cite{hu2020leveraging} & $65.1$ &$71.0$ \\
        &LaplacianShot~\cite{MasudZiko2020} & $73.5$ &$86.8$ \\
        &TIM~\cite{boudiaf2020transductive} &$75.8$ & $85.4$\\ 
        &$\alpha$-TIM~\cite{veilleux2021realistic} &$76.0$ & $87.8$\\ 
        &EASY 3$\times$ResNet12 \scriptsize{(ours)} & $\mathbf{78.46}$ & $\mathbf{87.85}$ \\
    \bottomrule
    \end{tabular}
    \label{tab:results_imbalanced_transductive_tieredImageNet}
\end{table}

\begin{table}[!htbp]
    \caption{1-shot and 5-shot accuracy of state-of-the-art methods and proposed solution on \textbf{CUB-FS} in \textbf{imbalanced transductive} setting.}
    \centering
    \small
    \begin{tabular}{p{0.2cm}p{3.7cm}>{\centering\arraybackslash}p{1.6cm}>{\centering\arraybackslash}p{1.6cm}}
    \toprule
        & Method & 1-shot & 5-shot \\
        \midrule
         \multirow{4}{*}{\rotatebox[origin=c]{90}{$36M$}
        $\left\{\vphantom{\begin{tabular}{c}~\\~\\~\\~\\~\\~\\\end{tabular}}\right.$}
        &PT+MAP~\cite{hu2020leveraging}  &$65.1$ & $71.3$\\ 
        &Entropy-min~\cite{dhillon2019baseline}   &$67.5$ & $82.9$\\
        &LaplacianShot~\cite{MasudZiko2020} &$73.7$ & $87.7$\\ 
        &TIM~\cite{boudiaf2020transductive}  &$74.8$ & $86.9$\\ 
        &$\alpha$-TIM~\cite{veilleux2021realistic} &$75.7$ & $89.8$\\ 
        &EASY 3$\times$ResNet12 \scriptsize{(ours)} & $\mathbf{83.63}$ & $\mathbf{92.35}$\\
    \bottomrule
    \end{tabular}
    \label{tab:results_imbalanced_transductive_cubfs}
\end{table}
\section{Additional ablation studies}

\subsection{Influence of the temperature in the transductive setting}
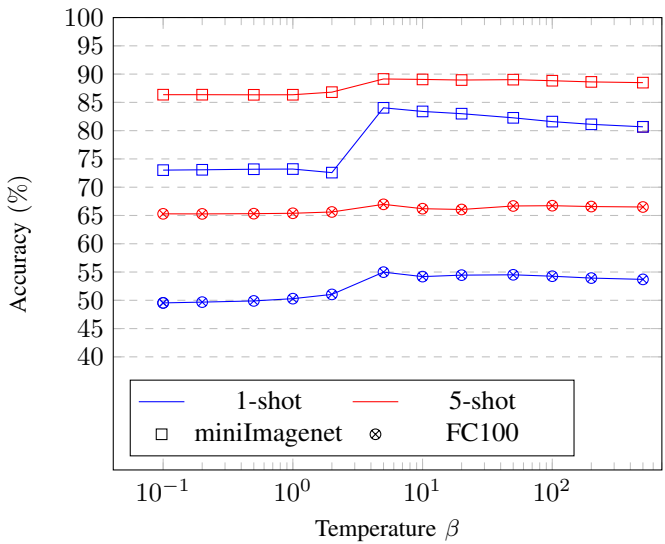
\begin{figure}[!htbp]
\begin{tikzpicture}
\centering
\begin{axis}[
    xlabel={\small{Temperature $\beta$}},
    ylabel={\small{Accuracy $(\%)$}},
    xmin=0, xmax=700,
    ymin=20, ymax=100,
    height=7.58cm,
    width=8.8cm,
    xtick={0, 0.1, 1, 10,100,1000},
    ytick={40, 45, 50, 55, 60, 65,70,75,80,85,90, 95, 100},
    legend pos=south west,
    legend columns=2, 
    legend style={minimum width=2.2cm},
    ymajorgrids=true,
    grid style=dashed,
    xmode=log,
]
\addplot[
    color=blue,
    mark = none,
    ]
    coordinates { %miniImagenet
   (500,80.68)
    };
    \addlegendentry{1-shot};
\addplot[
    color=red,
    mark = none,
    ]
    coordinates { %miniImagenet
   (0.1,86.36)
    };

    \addlegendentry{5-shot};    
\addplot[
scatter, 
only marks, 
    color=black,
    mark = square,
    ]
    coordinates { %miniImagenet
   (500,80.68)
    };

    \addlegendentry{miniImagenet};    
\addplot[
scatter, 
only marks, 
    color=black,
    mark = otimes,
    ]
    coordinates { %FC100
   (0.1,49.54)
    };

    \addlegendentry{FC100};    
    
\addplot[
    color=blue,
    mark=square,
    ]
    coordinates { %miniImagenet
    (0.1,73.02)(0.2,73.08)(0.5,73.18)(1,73.21)(2,72.58)(5,84.04)(10,83.40)(20,83.00)(50,82.27)(100,81.60)(200,81.12)(500,80.68)
    };

\addplot[
    color=red,
    mark=square,
    ]
    coordinates {
    (0.1,86.36)(0.2,86.36)(0.5,86.34)(1,86.35)(2,86.81)(5,89.14)(10,89.06)(20,88.95)(50,89.02)(100,88.83)(200,88.62)(500,88.49)
    };
%\addlegendentry{5-shot miniImagenet};

\addplot[ % FC100 1-shot
    color=blue,
    mark=otimes,
    ]
    coordinates {
    (0.1,49.54)(0.2,49.68)(0.5,49.89)(1,50.29)(2,51.04)(5,54.98)(10,54.19)(20,54.45)(50,54.51)(100,54.26)(200,53.93)(500,53.70)
    };    
%\addlegendentry{1-shot FC100};

\addplot[ % FC100 5-shot
    color=red,
    mark=otimes,
    ]
    coordinates {
    (0.1,65.29)(0.2,65.28)(0.5,65.32)(1,65.39)(2,65.62)(5,66.97)(10,66.20)(20,66.06)(50,66.68)(100,66.73)(200,66.58)(500,66.50)
    };    
%\addlegendentry{5-shot FC100};

    %\legend{1-shot, 5-shot}
  
\end{axis}
\end{tikzpicture}

\caption{Ablation study of Temperature of the soft K-means used in the transductive setting. We perform $10^5$ runs for each value of $\beta$.} \label{fig:ablation_temperature}
\end{figure}

In Figure \ref{fig:ablation_temperature}, we show how different values of the temperature $\beta$ of the soft K-means influence the performance of our model. We observe that $\beta=5$ seems to lead to the best results on the two considered datasets, which is why we chose this value in our other experiments. Note that we use three ResNet12 with 30 augmented samples in this experiment.

\subsection{Influence of the number of crops}
\begin{figure*}[!htbp] % Samples ablation
\begin{center}
\begin{tikzpicture}
\centering
\begin{axis}[
    xlabel={\small{Number of crops $\ell$}},
    ylabel={\small{Accuracy $(\%)$}},
     height=10cm,
     width=18cm,
    xmin=0, xmax=700,
    ymin=22, ymax=100,
    xtick={0, 0.1, 1, 10,100,1000},
    ytick={40, 45, 50, 55, 60, 65,70,75,80,85,90, 95, 100},
    legend pos=south west,
    legend columns=2, 
    legend style={minimum width=2.5cm},
    ymajorgrids=true,
    grid style=dashed,
    xmode=log,
]

\addplot[
    color=blue,
    mark = none,
    ]
    coordinates { %1-shot
   (1,68.29)
    };
    \addlegendentry{1-shot};

\addplot[
    color=red,
    mark = none,
    ]
    coordinates { %5-shot
    (1,83.84)
    };

    \addlegendentry{5-shot};

\addplot[
    color=black,
    mark = none,
    dashed,
    ]
    coordinates { %Inductive
   (1,68.29)
    };
\addlegendentry{inductive};

\addplot[
    color=black,
    mark = none,
    ]
    coordinates { %Inductive
   (1,68.29)
    };

    \addlegendentry{transductive};
    
\addplot[
scatter, 
only marks, 
    color=black,
    mark = square,
    ]
    coordinates { %miniImagenet
   (1,83.84)
    };
    \addlegendentry{miniImagenet};    
\addplot[
scatter, 
only marks, 
    color=black,
    mark = otimes,
    ]
    coordinates { %FC100
    (1,45.78)
    };

    \addlegendentry{FC100};    
    
\addplot[
    color=blue,
    mark=square,
    dashed, 
    mark options=solid,
    ]
    coordinates { %miniImagenet Inductive 1-shot
     (1,68.29)(2,66.72)(3,68.04)(5,69.47)(10,70.12)(20,70.70)(30,70.81)(50,70.70)(100,70.84)(200,71.00)(300,70.93)(500,71.29)
    };

\addplot[
    color=red,
    mark=square,
    dashed, 
    mark options=solid,
    ]
    coordinates { %miniImagenet Inductive 5-shot
    (1,83.84)(2,82.69)(3,83.58)(5,84.71)(10,85.46)(20,85.86)(30,85.79)(50,85.97)(100,85.94)(200,86.03)(300,86.07)(500,86.18)
    };
    
\addplot[
    color=blue,
    mark=square,
    ]
    coordinates { %miniImagenet Transductive 1-shot
     (1,80.13)(2,77.33)(3,79.12)(5,80.26)(10,81.07)(20,81.33)(30,81.56)(50,81.37)(100,81.63)(200,81.80)(300,82.08)(500,81.96)
    };
\addplot[
    color=red,
    mark=square,
    ]
    coordinates { %miniImagenet Transductive 5-shot
    (1,86.91)(2,85.59)(3,86.19)(5,87.35)(10,87.80)(20,88.23)(30,88.40)(50,88.41)(100,88.41)(200,88.41)(300,88.47)(500,88.44)
    };

%\addlegendentry{5-shot miniImagenet};

\addplot[ % FC100 1-shot Inductive
    color=blue,
    mark=otimes,
    dashed, 
    mark options=solid,
    ]
    coordinates { %FC100 Inductive 1-shot
    (1,45.78)(2,44.19)(3,45.03)(5,45.84)(10,46.24)(20,46.50)(30,46.64)(50,46.67)(100,46.72)(200,46.76)(300,46.71)(500,46.63)
    };    
\addplot[ % FC100 5-shot Inductive
    color=red,
    mark=otimes,
    dashed, 
    mark options=solid,
    ]
    coordinates { %FC100 Inductive 5shot
    (1,62.92)(2,61.10)(3,62.49)(5,63.24)(10,63.69)(20,64.21)(30,64.12)(50,64.26)(100,64.20)(200,64.35)(300,64.34)(500,64.29)
    };    

\addplot[ % FC100 1-shot Transductive
    color=blue,
    mark=otimes,
    ]
    coordinates { %FC100 Transductive 1-shot
    (1,51.86)(2,49.61)(3,50.86)(5,51.71)(10,52.42)(20,52.45)(30,52.29)(50,52.67)(100,52.80)(200,52.73)(300,52.72)(500,52.89)
    };    

\addplot[ % FC100 5-shot Transductive
    color=red,
    mark=otimes,
    ]
    coordinates { %FC100 transductive 5shot
    (1,64.67)(2,63.00)(3,64.11)(5,64.94)(10,65.34)(20,65.63)(30,65.60)(50,65.57)(100,65.67)(200,65.65)(300,65.59)(500,65.69)
    };    
\end{axis}
\end{tikzpicture}
\caption{Ablation study of Augmented Samples, we perform $10^5$ runs for each value of $\ell$.} \label{fig:ablation_crops}
\end{center}
\end{figure*}
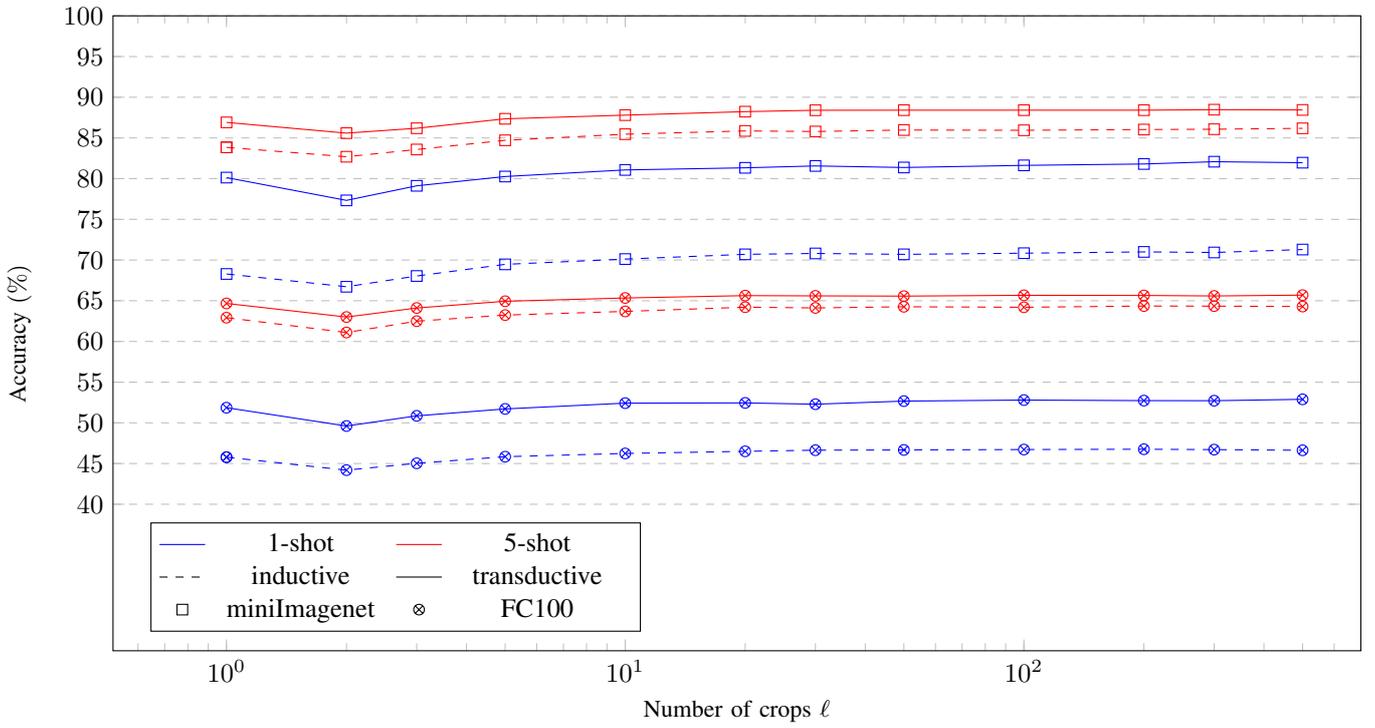

In Figure \ref{fig:ablation_crops}, we show how the performance of our model is influenced by the number of crops $\ell$ used during Augmented Sampling (AS). When using $\ell=1$, we report the performance of the method using no crops but a global reshape instead. We observe that the performance keeps increasing as long as the number of crops used is increased, except for a small drop of performance when switching from a global reshape to crops --this drop can easily be explained as crops are likely to miss the object of interest--. However, the computational time to generate the crops also increases linearly. Therefore, we use $\ell=30$ as a trade-off between performance and time complexity. Here, we use a single ResNet12 for our experiments.

\subsection{Influence of the number of backbones}
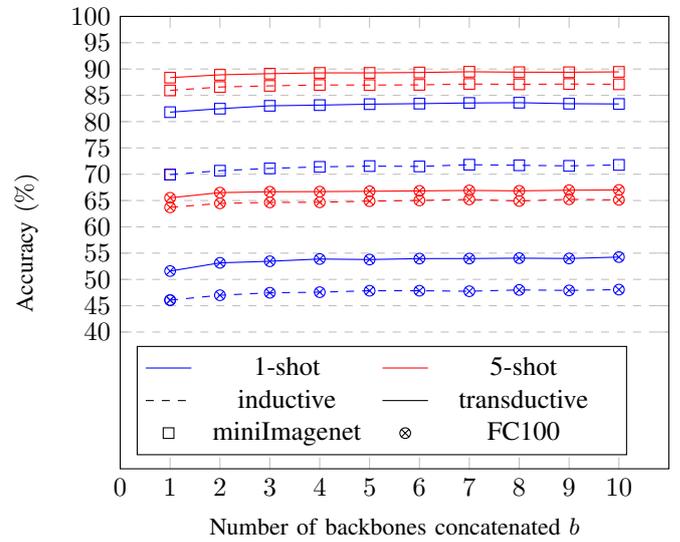
\begin{figure}[!htbp] % Samples ablation
\begin{center}
\begin{tikzpicture}
\centering
\begin{axis}[
    xlabel={\small{Number of backbones concatenated $b$}},
    ylabel={\small{Accuracy $(\%)$}},
    xmin=0, xmax=11,
    ymin=14, ymax=100,
    height=7.58cm,
    width=8.8cm,
    xtick={0, 1, 2, 3, 4, 5, 6, 7, 8, 9, 10},
    ytick={40, 45, 50, 55, 60, 65,70,75,80,85,90, 95, 100},
    legend pos=south west,
    legend columns=2, 
    legend style={minimum width=2.5cm},
    ymajorgrids=true,
    grid style=dashed,
]

\addplot[
    color=blue,
    mark = none,
    ]
    coordinates { %1-shot
   (1,69.93)
    };
    \addlegendentry{1-shot};

\addplot[
    color=red,
    mark = none,
    ]
    coordinates { %5-shot
    (1,83.84)
    };

    \addlegendentry{5-shot};

\addplot[
    color=black,
    mark = none,
    dashed,
    ]
    coordinates { %Inductive
   (1,68.29)
    };
\addlegendentry{inductive};

\addplot[
    color=black,
    mark = none,
    ]
    coordinates { %Inductive
   (1,69.93)
    };

    \addlegendentry{transductive};
    
\addplot[
scatter, 
only marks, 
    color=black,
    mark = square,
    ]
    coordinates { %miniImagenet
   (1,69.93)
    };
    \addlegendentry{miniImagenet};    
\addplot[
scatter, 
only marks, 
    color=black,
    mark = otimes,
    ]
    coordinates { %FC100
    (1,46.06)
    };

    \addlegendentry{FC100};    

\addplot[
    color=blue,
    mark=square,
    dashed, 
    mark options=solid,
    ]
    coordinates { %miniImagenet Inductive 1-shot
     (1,69.93)(2,70.67)(3,71.11)(4,71.40)(5,71.55)(6,71.48)(7,71.81)(8,71.70)(9,71.59)(10,71.77 )
    };

\addplot[
    color=red,
    mark=square,
    dashed, 
    mark options=solid,
    ]
    coordinates { %miniImagenet Inductive 5-shot
    (1,85.93)(2,86.58)(3,86.80)(4,86.97)(5,86.95)(6,86.99)(7,87.15)(8,87.09)(9,87.15)(10,87.10)
    };
\addplot[
    color=blue,
    mark=square,
    ]
    coordinates { %miniImagenet Transductive 1-shot
     (1,81.80)(2,82.46)(3,83.01)(4,83.14)(5,83.32)(6,83.42)(7,83.55)(8,83.60)(9,83.41)(10,83.36)
    };

\addplot[
    color=red,
    mark=square,
    ]
    coordinates { %miniImagenet Transductive 5-shot
     (1,88.36)(2,88.90)(3,89.12)(4,89.27)(5,89.27)(6,89.34)(7,89.48)(8,89.40)(9,89.39)(10,89.46)
    };

%\addlegendentry{5-shot miniImagenet};

\addplot[ % FC100 1-shot Inductive
    color=blue,
    mark=otimes,
    dashed, 
    mark options=solid,
    ]
    coordinates { %FC100 Inductive 1-shot
     (1,46.06)(2,46.99)(3,47.46)(4,47.57)(5,47.85)(6,47.85)(7,47.75)(8,48.00)(9,47.91)(10,48.06)
    };    

\addplot[ % FC100 5-shot Inductive
    color=red,
    mark=otimes,
    dashed, 
    mark options=solid,
    ]
    coordinates { %FC100 Inductive 5shot
     (1,63.70)(2,64.46)(3,64.64)(4,64.66)(5,64.88)(6,65.01)(7,65.21)(8,64.91)(9,65.23)(10,65.12)
    };    

\addplot[ % FC100 1-shot Transductive
    color=blue,
    mark=otimes,
    ]
    coordinates { %FC100 Transductive 1-shot
     (1,51.59)(2,53.15)(3,53.45)(4,53.88)(5,53.78)(6,53.96)(7,53.97)(8,54.04)(9,53.98)(10,54.25)
    };    

\addplot[ % FC100 5-shot Transductive
    color=red,
    mark=otimes,
    ]
    coordinates { %FC100 transductive 5shot
       (1,65.52)(2,66.50)(3,66.67)(4,66.69)(5,66.76)(6,66.83)(7,66.92)(8,66.84)(9,66.96)(10,67.01)
    };  

\end{axis}
\end{tikzpicture}
\caption{Ablation study of the number of backbones, we perform $10^5$ runs for each value of $b$.} \label{fig:ablation_backbones}
\end{center}
\end{figure}

In Figure \ref{fig:ablation_backbones}, we show how the performance of our model is influenced by the number of backbones $b$ used during the Ensemble step (E). The performance increases steadily with a strong diminishing return. We use 30 augmented samples in this experiment.

